\documentclass[a4paper, 10pt, conference]{ieeeconf}

\IEEEoverridecommandlockouts  

\makeatletter
\let\NAT@parse\undefined
\makeatother

\usepackage[dvipsnames]{xcolor}
\usepackage[colorinlistoftodos]{todonotes}

\newcommand*\linkcolours{ForestGreen}
\usepackage{times}
\usepackage{graphicx}
\usepackage{amssymb}
\usepackage{gensymb}
\usepackage{amsmath}
\usepackage{breakurl}

\usepackage{url,hyperref}
\hypersetup{
colorlinks,
linkcolor=\linkcolours,
citecolor=\linkcolours,
filecolor=\linkcolours,
urlcolor=\linkcolours}

\usepackage{algorithm}
\usepackage{algorithmic}

\usepackage[labelfont={bf},font=small]{caption}
\usepackage[none]{hyphenat}

\usepackage{mathtools, cuted}

\usepackage[noadjust, nobreak]{cite}

\usepackage{tabularx}
\usepackage{amsmath}

\usepackage{float}

\usepackage{pifont}

\newcolumntype{Y}{>{\centering\arraybackslash}X}

\usepackage[]{placeins}

\usepackage{placeins}

\usepackage{tikz}

\usepackage[framemethod=tikz]{mdframed}

\usepackage{afterpage}

\usepackage{stfloats}

\usepackage{atbegshi}
\newcommand{\handlethispage}{}
\newcommand{\discardpagesfromhere}{\let\handlethispage\AtBeginShipoutDiscard}
\newcommand{\keeppagesfromhere}{\let\handlethispage\relax}
\AtBeginShipout{\handlethispage}

\usepackage{comment}
\usepackage{multirow}
\usepackage{booktabs}
\usepackage{siunitx}
\usepackage{multicol}
\usepackage{graphicx}
\usepackage{subfigure}

\usepackage[numbers]{natbib}

\setcitestyle{open={[},close={]}}

\title{\LARGE \bf
Simplifying Two-Stage Detectors for On-Device Inference in Remote Sensing}

\author{Jaemin Kang$^{1}$ Hoeseok Yang$^{2}$ Hyungshin Kim$^{3}$%
\thanks{$^{1}$ Jaemin, Email: jaemin.k96@o.cnu.ac.kr}%
\thanks{$^{2}$ Hoeseok, Email: hoeseok.yang@scu.edu}%
\thanks{$^{3}$ Hyungshin, Email: hyungshin@cnu.ac.kr}%
}

\begin{document}

\maketitle
\thispagestyle{empty}
\pagestyle{empty}

\begin{abstract}

Deep learning has been successfully applied to object detection from remotely sensed images. Images are typically processed on the ground rather than on-board due to the computation power of the ground system. Such offloaded processing causes delays in acquiring target mission information, which hinders its application to real-time use cases. For on-device object detection, researches have been conducted on designing efficient detectors or model compression to reduce inference latency. However, highly accurate two-stage detectors still need further exploitation for acceleration. In this paper, we propose a model simplification method for two-stage object detectors. Instead of constructing a general feature pyramid, we utilize only one feature extraction in the two-stage detector. To compensate for the accuracy drop, we apply a high pass filter to the RPN's score map. Our approach is applicable to any two-stage detector using a feature pyramid network.
In the experiments with state-of-the-art two-stage detectors such as ReDet, Oriented-RCNN, and LSKNet, our method reduced computation costs upto 61.2\% with the accuracy loss within 2.1\% on the DOTAv1.5 dataset. Source code will be released.

\end{abstract}

\section{Introduction}
\label{sec:intro}
Deep learning-based object detection is widely used for remotely sensed images taken from UAVs or satellites. Currently, object detection is performed on the ground once images are transferred to the ground facility. On the ground, powerful GPU clusters can be used to run highly accurate and hence complex deep learning models. However, considering end-to-end delay from the acquisition of the image to the inference, large delay from a few minutes upto a few days incurs until the image is arrived at the ground facility~\cite{qi2018board,shen2023board,ghiglione2022opportunities}. To curtail this delay and to process in real time, on-board AI is gaining attention. 
With on-board AI, we run deep learning inference on-board UAVs or satellites. 

There are difficulties in performing on-board inference of high-precision object detection models due to the constrained computation capability and power budget. State-of-the-art object detection deep learning models for remote sensing imagery have a two-stage structure with an Oriented RPN head~\cite{li2303large,yu2023spatial,xie2021oriented}. Providing real-time performance is difficult due to the computational complexity of high-accuracy object detection models. Some one-stage detectors provide real-time performance, but they have lower accuracy compared to other high-accuracy detectors~\cite{zhao2019robust}. The use of low-accuracy models poses a significant risk when their applications are in mission critical domains. As a result, researchers are working to create more effective models aiming for great accuracy but also very fast. This paper is on simplifying highly accurate two-stage detectors while maintaining accuracy.

\begin{figure}[t]
    \centering
    \includegraphics[width=1\linewidth]{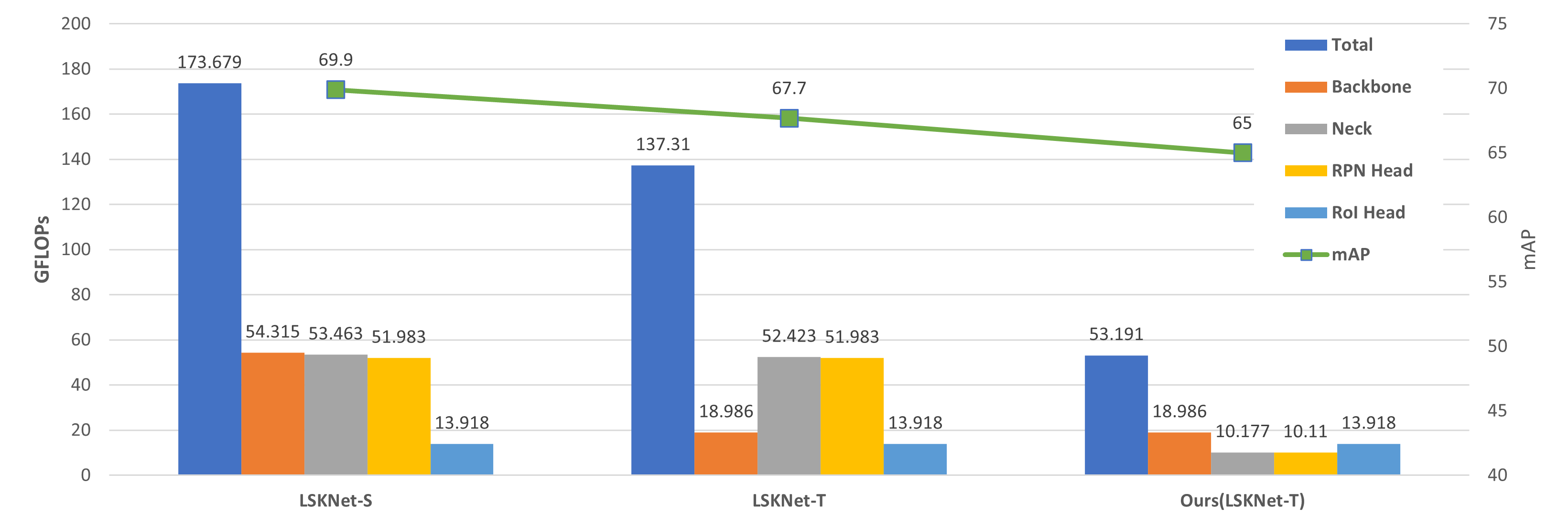}
    \caption{ Computation workload analysis of a two-stage detector, LSKNet~\cite{li2303large}. 
    The size of the input image is 1024 pixels by 1024 pixels. LSKNet-T is created by reducing the backbone of the LSTNet-S. Graph on the right shows our method is effective to the regression part of the detector.}
    \label{fig:Flops of two-stage detectors}
\end{figure}

Several studies have proposed various methods to lighten and speed up the model without compromising its accuracy. Pruning is a technique that lightens a model by deleting redundant weights~\cite{lyu2023survey}. They remove weights in structural modules and create a faster model than the original. However, when applying the pruning technique to a two-stage detector, it is generally applicable to the backbone. Thus, even after pruning, there are still many FLOPs left for regression. Figure~\ref{fig:Flops of two-stage detectors} shows the computational load for each component of a two stage detector. The figure shows the impact of pruning on a two stage detector LSKNet~\cite{li2303large}, which demonstrates high accuracy with low FLOPs. In LSKNet-S (Small model), the backbone, neck, and RPN head exhibit similar FLOPs. From the LSKNet-S, compressed model LSKNet-T (Tiny model) is created with reduced number of parameters. In LSKNet-T, the FLOPs of the backbone are reduced by 3X than the LSKNet-S, but those of the neck and RPN head remain the same. Computations in the regression part should be further exploited to get extra speed up. The right most chart of Figure~\ref{fig:Flops of two-stage detectors} shows our compression affects on the regression parts of the detector.

YOLOF~\cite{chen2021you} is the study on simplifying a one-stage detector with single feature. It demonstrates single feature is good enough to detect objects of various size. It addressed the regression overhead problem by employing dilated convolution and uniform matching with the single feature.
However, it still remains as a challenge to maintain an acceptable accuracy level for small objects~\cite{yi2023point2rbox}.
From the DOTA-v1.0 dataset~\cite{xia2018dota}, which has many small objects from remotely sensed images, YOLOF only achieves an accuracy of 66.54\%, whereas the RetinaNet~\cite{lin2017focal} model achieves an accuracy of 68.69\%. Note that a two-stage detector, Oriented R-CNN~\cite{xie2021oriented}, could achieve an accuracy of 75.87\% on the same dataset.

\begin{figure*}[t]
    \centering
    \subfigure[General Two-stage detector]{\includegraphics[width=0.49\textwidth]{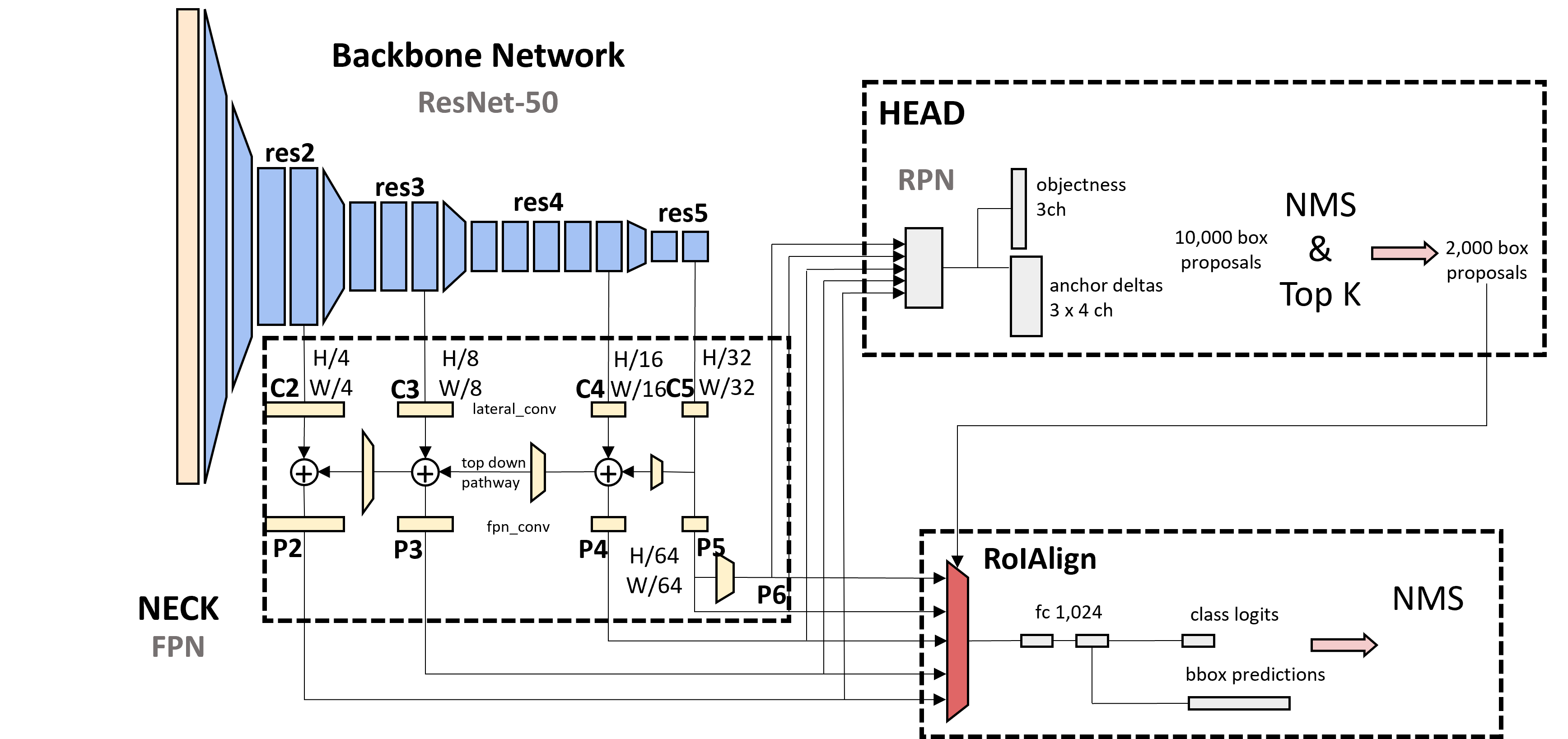}\label{fig:fig2_a}}
    \hfill
    \subfigure[Two-stage detector adopting our methods]{\includegraphics[width=0.49\textwidth]{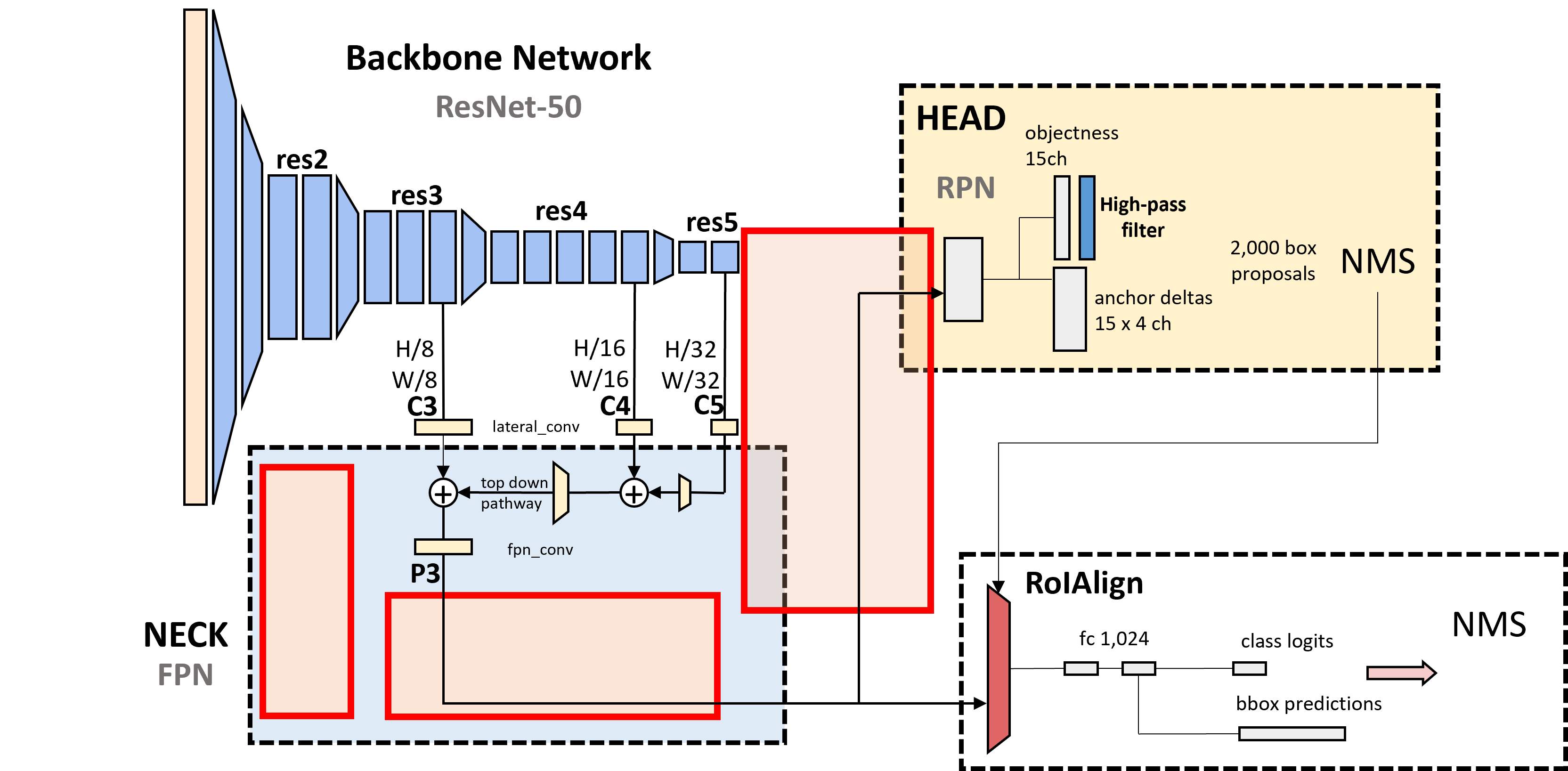}\label{fig:fig2_b}}
    \caption{A two-stage detector has been implemented using our approach. Layers needed for constructing the feature pyramid have been removed. Furthermore, because both RPN and RoIAlign utilize a single feature, operations that are not included in FLOPs are also reduced.}
    \label{fig:subfigures}
\end{figure*}

In this paper, we propose a model compression method by simplifying regression part of two-stage detectors. Figure~\ref{fig:fig2_b} illustrates the schematic representation of the model implemented using our approach. We perform regression using only single feature without constructing a feature pyramid. By removing the feature pyramid structure, we reduce the computational load of the RPN, NMS, and RoIAlign for regression. Simple removal of the feature pyramid incurs large accuracy drop. 
We have to overcome two challenges to maintain the baseline accuracy when removing the feature pyramid. One is that the IoU (Intersection over Union) between the anchor and the object decreases than the positive anchor threshold. This results in low accuracy since the detector is unable to learn small objects. We choose features that exceed a threshold of positive anchors with small objects, regardless of the anchor position. The other is that the RoIs are generated focusing on large objects. Many RoI are generated focusing on relatively easy-to-detect large objects. When several objects appear in an image, small objects tend to be undetected. As a solution to this challenge, we design a high-pass filter to focus on the RoIs of small objects.

Our approach is advantageous because it is applicable to any two-stage detector with a feature pyramid network. In the experiments with state-of-the-art two-stage detectors such as ReDet~\cite{han2021redet}, Oriented-RCNN~\cite{xie2021oriented}, and LSKNet~\cite{li2303large}, STDNet~\cite{yu2023spatial}, our method reduced computation costs upto 55.6\% with the accuracy loss within 2.7\% on the DOTAv1.5 dataset~\cite{xia2018dota}.

\section{RELATED WORK}
\label{sec:rel}
\subsection{Remote sensing object detectors}
Researches have been conducted to efficiently regress HBB~\cite{zhou2021probabilistic,terven2023comprehensive,dai2021dynamic}.
The approaches have shown excellent efficiency in ground-based detection but have challenges when applied to OBB.
Objects such as buildings, vehicles, or natural features in remote sensing images have a larger aspect ratio compared to images captured from the ground~\cite{xia2018dota}. Moreover, objects in remote sensing images may not align perfectly with horizontal or vertical axes, resulting in arbitrary orientations. Traditional horizontal bounding boxes (HBB) may not accurately capture the scope of the object of interest and can include a lot of surrounding image noise when performing object regression. To address these issues, oriented bounding boxes (OBB) are used. The rotated bounding box aligns with the object's orientation, allowing for more precise positioning.

The simplest method to convert an HBB detector to an OBB is by regressing the rotation angle along with the HBB.
However, this method has relatively lower accuracy compared to other methods. Researchers are conducting a study to efficiently regress OBB.
Using the rotational anchors ~\cite{ma2018arbitrary} is straightforward for regressing OBB, but it requires an extra computational load compared to HBB. Ding et al,~\cite{ding2019learning} proposed the RoI transformer. It regresses OBB using fully connected layers that makes the network heavy and complex. Xie, Xingxing, et al,~\cite{xie2021oriented}. proposed Oriented-RPN, which minimizes additional computations regressing OBB without parameters for anchor and angle. It is similar to HBB's RPN but regresses OBB throughout decoding. Lyu, Chengqi, et al, proposed RTMDet ~\cite{lyu2022rtmdet} for real-time object detectors. It has high accuracy and fast inference speed with an anchor-free, one-stage detector. However, in situations where multi-scale testing cannot be conducted for real-time purposes. The STD~\cite{yu2023spatial} and LSKNet~\cite{li2303large} models in the DOTAv1.0~\cite{xia2018dota} dataset have achieved state-of-the-art accuracy by using an Oriented RPN head in a two-stage detector. We propose a model simplification method applicable to two-stage detectors to achieve real-time performance on high-accuracy models.

\subsection{Detectors using single feature}

The early object detector models that employed deep learning, such as the R-CNN series~\cite{girshick2014rich,girshick2015fast,ren2015faster}, utilized single feature.
In Faster R-CNN~\cite{ren2015faster}, the use of anchors with different sizes was introduced to detect objects of varying sizes. One-stage detectors like YOLO~\cite{redmon2016you} and YOLOv2~\cite{redmon2017yolo9000} also used only single feature.
In YOLOv2, anchors were extracted based on the dataset to enhance accuracy, but the absence of multi-scale capabilities still resulted in accuracy limitations.
FPN~\cite{lin2017feature} network is used to achieve high accuracy in subsequent detectors. However, when using FPN, doing regression and classification on each feature that makes up the feature pyramid requires a significant amount of computational and memory resources.

Research exists that aims to achieve efficient detection using just a single feature without constructing a feature pyramid. YOLOF~\cite{chen2021you} used only the C5 feature with the dilated encoder and uniform matching. Zhou et al.~\cite{zhou2019objects} proposed CenterNet, which is an anchor-free object detector.
It is a detector developed not for efficiency but using a single feature.
CenterNet achieved high accuracy while being efficient by using the P2 feature.
They have comparable accuracy to FPN-based detectors on the COCO dataset. 
But they reveal a decrease in accuracy when detecting small objects in aerial datasets.
We explain the difficulty in detecting small objects when using just one feature in a detector. We propose a method of using just one feature while minimizing accuracy loss. Our approach maintains accuracy in detecting small objects while maintaining resource efficiency with a single feature.

\section{Methods}
\label{sec:Methods}
In this paper, we propose a method that performs regression using just only one feature from the existing detector.
Our method involves the selection of one feature from the features that constitute the feature pyramid at the neck structure.
The selected feature is used to find objects having various sizes.
We attach the anchors from the removed features to the selected feature.
We achieved this task by increasing the number of anchors used in the feature. It results in an increase in the channel parameter of the convolution layer in the RPN.
However, the removed features do not undergo the RPN head. This results in a significant reduction in FLOPs.
In this section, we explain our approach for selecting a single feature to accelerate inference while maintaining accuracy.

\begin{figure}[t]
    \centering
    \includegraphics[width=1\linewidth]{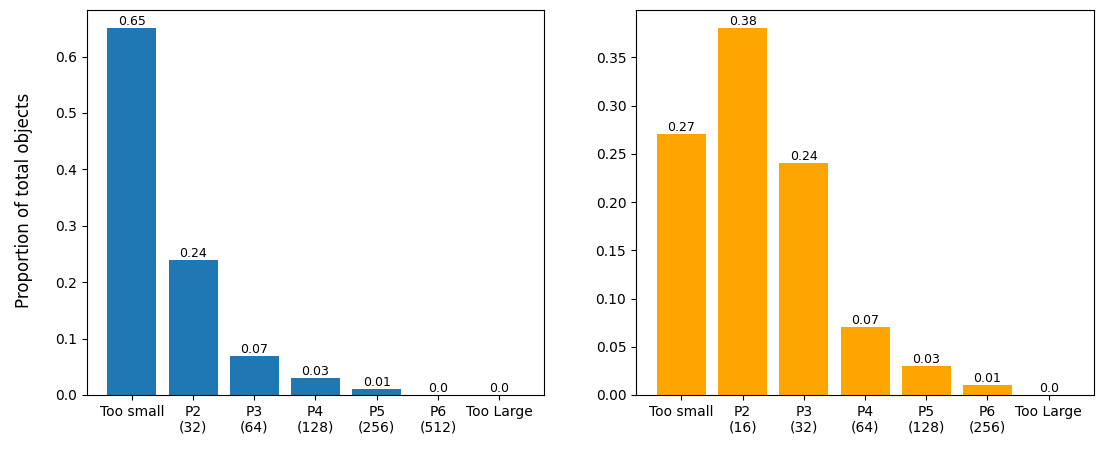}
    \caption{ Ratio of matched anchors with objects in each feature. The graph shows proportion of objects in the validation dataset of DOTAv1.5 with an IoU of 0.5 or higher as the matched anchor. The left graph is the result from the original model with anchor sizes of 32, 64, 128, 256, and 512. The right graph is the result from the modified anchors of sizes 16, 32, 64, 128, and 256.  }
    \label{fig:object distribution}
\end{figure}

\subsection{Adjusting the anchor size}
\label{sec:3.1}
The performance of an anchor-based detector is tied to the size of the anchor.
Because the anchor serves as a reference point for the detector, its size influences the detector's ability to accurately identify and classify objects.
Unlike two-stage detectors, the YOLO series takes a different approach by extracting anchors based on the dataset.
This allows the YOLO series to adapt to the specific characteristics of the dataset, potentially improving detection accuracy.
However, two-stage detectors typically utilize anchor sizes that were used in early research.
This early research primarily focused on detecting animals and objects on the ground, which are significantly different from the objects found in remote sensing datasets.
As a result, anchor sizes do not match well with remote sensing datasets, leading to potential inaccuracies in detection.

Figure \ref{fig:object distribution} shows that the pre-generated anchors previously used do not align with the remote sensing dataset. The left graph shows the matched ratio with original anchor sizes, while the right graph represents ratio with the adjusted anchor sizes. In the DOTAv1.5 dataset, it is shown that 65\% of objects are smaller than the objects detected by the P2 feature in the original detector. This suggests that the original anchor sizes may not be optimal for this dataset.
To address this issue, we adjust the size of anchors used by the detector to match that of the objects. The right graph in Figure \ref{fig:object distribution} displays the number of objects detectable by the adjusted anchor size in each feature. Despite these adjustments, 27\% of objects are still smaller than the detectable size at P2. This indicates that further adjustments may be necessary to improve detection accuracy.
However, if we continue reducing the anchor size further, the anchor size becomes too small to cover object area for detection. In the next section, we will explain the relationship between features and anchors. Based on this explanation, we modify anchor sizes to enhance the detection of small objects in the DOTAv1.5 dataset.
We choose anchor sizes that are multiples of the stride, considering the anchor's stride. Since using smaller anchors necessitates significant computer processing, we have selected a minimum anchor size of 16 for efficient detection. Since the scale of the anchor was divided based on an IoU of 0.5, we maintain the scale as a multiple of 2. Finally, anchor sizes of 16, 32, 64, 128, and 256 are used instead of the previous sizes of 32, 64, 128, 256, and 512.

\subsection{Selecting a single feature for inference acceleration}

Detectors utilize anchors for efficient object detection training.
Training on randomly generated boxes results in longer learning times. Thus, the main approach is to use anchors as proposals in the training of the detector, enabling the detector to quickly learn object regression. Calculating the IoU between anchors and objects to choose anchors with high IoU as proposals for the detector, which enhances the efficiency of the learning process.

The downsample factor of the feature determines the density of the anchor.
Anchors are placed at each pixel of downsampled features. When IoU between objects and anchors is calculated, features with a small down-sampling ratio have more number of pixels to compute due to their larger size. Conversely, features with a high down-sampling ratio have fewer pixels, thus requiring less memory and computation. This is particularly beneficial when the goal is to perform regression using a single feature. A higher downsample factor of the feature allows for constructing the detector with less computational workload, making the process more efficient.

\begin{figure}[t]
    \includegraphics[width=\linewidth]{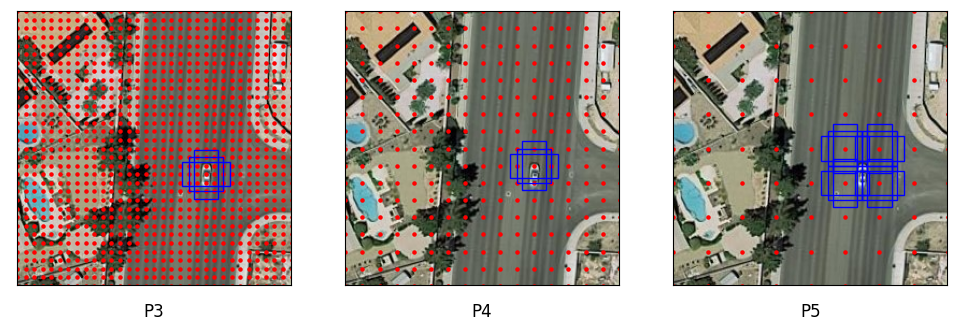}
    \caption{ Visualization of the anchor's stride based on the downscale factor of the features when the image size is 256 x 256. The red dot represents the location where the anchors appear in the feature. To detect a car, an anchor size of $32^2$ is sufficient, however it is not trained on the P5 feature due to low IoU. In P3 and P4 features, the IoU between the anchor and object exceeds the threshold for positive anchors.  }
    \label{fig:anchor visualization}
\end{figure}

However, selecting features with a high downsample factor significantly decreases the accuracy of the detector. The sparsity of anchors results in the IoU between small objects and anchors not surpassing the threshold for being selected as proposals during training. Figure~\ref{fig:anchor visualization} visualizes anchors and objects from features with three different downsampling ratios. Red dots indicate locations of anchors. The blue box represents an anchor. In the P3 feature, anchors have a high IoU with objects in multiple pixels. In the P4 feature, a higher IoU is achieved with fewer pixels used. However, because of the sparse anchors in the P5 feature, the IoU with the object does not exceed the threshold.

We aim to select features that can avoid detection failure even when the size of the anchor and the size of the object are the same.
For example, in Figure ~\ref{fig:anchor visualization}, if an object appears between anchors as in the P5 feature, it does not exceed the positive anchor threshold.
We observe that the IoU between anchors and objects is always higher than the positive anchor in features with a downsample factor less than the anchor size.
Thus, when constructing a detector having a single feature, it is better to use a downsampling factor less than the minimum size of the anchor. However, using a feature with a downsampling factor that is too small compared to the anchor poses another challenge. Using features with a very small downscale factor decreases accuracy. Noise occurs at the NMS stage after RoI generation. Furthermore, RoIs are focused on large objects, neglecting RoIs for small objects.

In our implementation, we use features with a downsample factor of 8, which is less than the minimum anchor size determined in the previous section~\ref{sec:3.1}, which was 16. We retain the P3 feature. We remove all other parts except for the layers needed to construct the P3 feature as shown in Figure~\ref{fig:fig2_b}.
Before using our method, the features of the feature pyramid were used as inputs for each RPN head. The detector extracts a number of RoIs from each feature map after the RPN head. Since we have kept just one feature, the number of feature used as input for the RPN head is reduced. Thus, there is a significant decrease in FLOPs at the RPN head. Furthermore, the number of selected RoIs is also significantly reduced.

\subsection{Applying high-pass filter}
When selecting RoIs in a two-stage detector, RoIs with high scores in the score map after the RPN head are chosen.
In a feature pyramid, a large number of RoIs are selected and then regression is performed. Additionally, features are divided according to the size of the item, therefore there is no issue with different-sized objects having different RoI scores.

However, when the detector performs regression using just one intermediate feature, the number of RoIs decreases significantly. Additionally, objects are not discriminated by size. Large objects that are relatively easy to learn get high scores and obtain scores from many pixels. On the other hand, small objects tend to have relatively low scores and acquire scores from a small number of pixels. When extracting RoIs based on the score, RoI for small objects can be neglected.

\begin{figure}[t]
        \centering
        \includegraphics[width=1\linewidth]{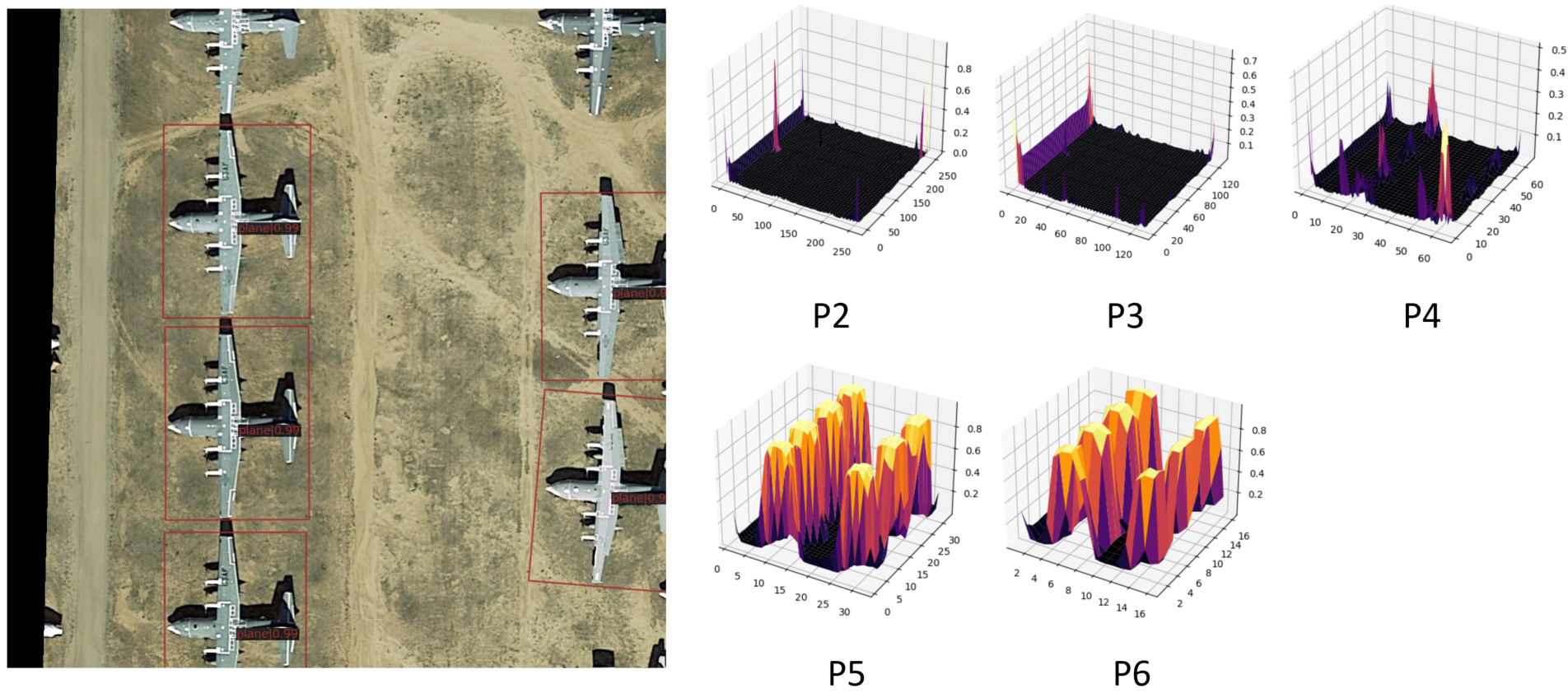}
        \caption{Classification score map after the picture has passed through the RPN in the detector. The score was measured in the Oriented R-CNN model. The score of graph is passed through a sigmoid function before extracting the RoI from the score.}
        \label{fig:rpn_score_map}
\end{figure}

\begin{figure}[t]
    \centering
    \includegraphics[width=1\linewidth]{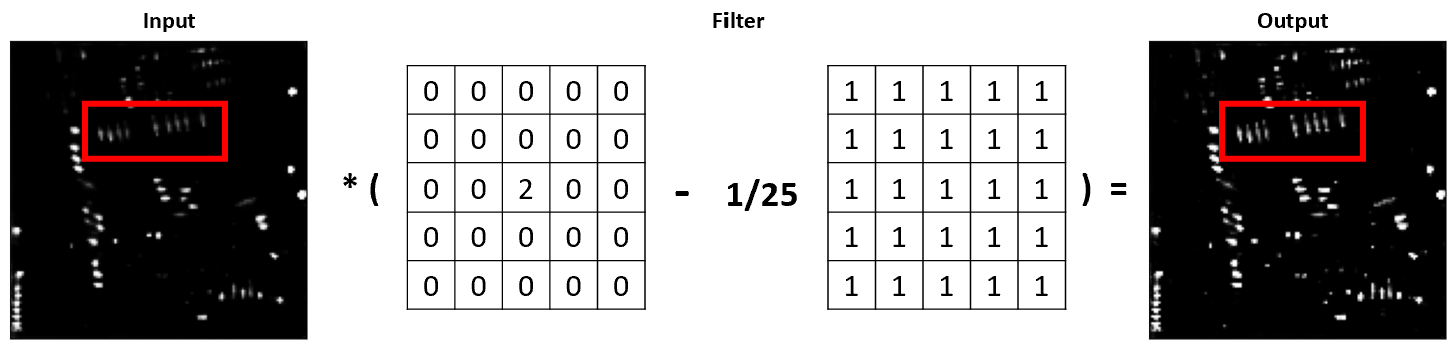}
    \caption{The Unsharp masking filter we used. The values around the center were set to 1, unlike the typical 5$\times$5 unsharp masking.}
    \label{fig:high-pass filter}
\end{figure}

Figure~\ref{fig:rpn_score_map} visualizes the score map for large objects. The object's score in the classification score map exhibits a bell shape curve. It is seen that several pixels get high scores for a single object. We attempt to narrow the variance of the score map for large objects and increase the peak for small objects.

A high-pass filter in image processing emphasizes high-frequency components and reduces low-frequency components in an image. It is commonly used to enhance edges or noise in images. We want to make the filter to sharpen for small objects and blur for large objects. We designed a 5$\times$5 filter that shows better performance than a 3$\times$3 filter. For a blur effect at the boundaries of a large object, we used the filter with uniform weights throughout rather than filters with increasing values towards the center.
Figure~\ref{fig:high-pass filter} displays the high-pass filter we used. objects. We get an accuracy improvement with fewer computations overhead by applying the filter.

\section{Experiments}
\label{sec:exp}

\subsection{Experiment setup}
Experiments are performed on the DOTA-v1.5~\cite{xia2018dota} dataset, a remote sensing dataset, to demonstrate the efficacy of our proposed method. 
DOTA-v1.5 uses the same images as DOTA-v1.0 but includes annotations for objects that are extremely tiny, less than 10 pixels in size.
It has 2,806 photos and 403,318 instances in total, classified into 16 object classes.
Since the DOTA dataset doesn't have labels for the test set, accuracy is evaluated with the validation set. 
The image sizes range from 800 to 4,000 pixels in both width and height.
The detector's input image commonly uses square dimensions.
We cropped the images to a 1024x1024 size with a stride of 200 pixels for training the detectors.

We have implemented our method to state-of-the-art two-stage models, such as ReDet~\cite{han2021redet}, Oriented-RCNN~\cite{xie2021oriented}, LSKNet~\cite{li2303large}, and STD~\cite{yu2023spatial}. 
Inference speed is measured in FPS on a NVIDIA 2080ti board. 

We used the mmrotate~\cite{zhou2022mmrotate} learning framework for our experiments. Models were trained using a single 3090 GPU. Hyperparameters other than learning rate remain the same as the existing parameters in the model training. We modified the learning rate values to enable training on a single GPU instead of several GPUs for the model training.

\begin{table*}[t]
\caption{Detection Accuracy(mAP) of each class of DOTA dataset. The ReDet model uses the ReResNet-50 backbone.
The Oriented R-CNN model employs ResNet-50.
The LSKNet-T model after applying our method has a decrease in accuracy when sent via the high-pass filter.
The models with an asterisk(*) indicate accuracy without being subjected to a high-pass filter.
Every model performed single scale training.} 
\label{tab:class_mAP}
\begin{center}
{\scriptsize
\begin{tabular}{l|c|c|c|c|c|c|c|c|c|c|c|c|c|c|c|c|c|c}
\hline
\hline

Model & PL & BD & BR & GTF & SV & LV & SH & TC & BC & ST & SBF & RA & HA & SP & HC & CC & mAP & FPS\\
\hline
ReDet~\cite{han2021redet} & 90.2 & 78.6 & 50.2 & \textbf{67.4} & \textbf{50.9} & 76.4 & 81.2 & 90.8 & 68.9 & 69.4 & 64.9 & 74.2 & 76.9 & 64.0 & 55.3 & 0.0 & \textbf{66.2} & 14.0 \\
Ours - ReDet & 90.2 & 78.5 & 48.3 & \textbf{63.4} & \textbf{55.9} & 74.9 & 81.2 & 90.7 & 68.4 & 71.1 & 64.4 & 67.6 & 77.1 & 63.7 & 60.7 & 0.0 & \textbf{66.0} & 22.0 \\
\hline
Oriented R-CNN~\cite{xie2021oriented}  & 90.0 & 76.6 & 55.1 & \textbf{74.0} & \textbf{50.3} & 76.6 & 88.3 & 90.8 & 72.4 & 62.2 & 57.9 & 72.5 & 75.1 & 65.3 & 51.1 & 0.4 & \textbf{66.2} & 19.8 \\
Ours - O-RCNN & 89.6 & 76.3 & 46.5 & \textbf{67.3} & \textbf{54.1} & 75.5 & 87.8 & 90.8 & 63.4 & 69.8 & 62.6 & 71.9 & 75.0 & 64.3 & 57.4 & 0.0 & \textbf{65.8} & 30.4 \\
\hline
LSKNet-T~\cite{li2303large}  & 89.4 & 83.6 & 51.2 & \textbf{78.4} & \textbf{50.5} & 76.0 & 88.4 & 90.6 & 75.2 & 68.9 & 65.1 & 72.4 & 74.2 & 64.2 & 54.0 & 0.1 & \textbf{67.7} & 22.3 \\
Ours - LSKNet-T & 89.0 & 76.7 & 46.6 & \textbf{68.7} & \textbf{53.3} & 74.7 & 86.9 & 90.6 & 67.4 & 69.6 & 61.3 & 65.9 & 68.9 & 63.3 & 54.4 & 0.0 & \textbf{65.1} & 36.4 \\
Ours - LSKNet-T$^*$  & 89.1 & 82.0 & 46.7 & \textbf{69.0} & \textbf{54.1} & 75.4 & 80.4 & 90.7 & 67.5 & 69.7 & 62.2 & 66.6 & 76.1 & 64.3 & 55.4 & 0.0 & \textbf{65.6} & 37.1 \\
\hline
LSKNet-S~\cite{li2303large}  & 89.9 & 85.6 & 54.8 & \textbf{79.5} & \textbf{50.9} & 77.4 & 89.7 & 90.7 & 75.0 & 69.6 & 72.9 & 74.2 & 76.3 & 66.7 & 64.8 & 0.0 & \textbf{69.9} & 18.2 \\
Ours - LSKNet-S  & 89.8 & 85.0 & 55.3 & \textbf{72.8} & \textbf{55.4} & 76.0 & 88.1 & 90.8 & 74.4 & 70.6 & 71.6 & 67.3 & 76.4 & 63.0 & 60.3 & 0.0 & \textbf{68.5} & 26.3 \\
\hline
STD-O HiViT-B~\cite{yu2023spatial}  & 90.2 & 85.6 & 58.6 & \textbf{74.2} &\textbf{50.6} & 78.1 & 89.5 & 90.8 & 65.7 & 70.2 & 67.8 & 75.4 & 76.6 & 67.0 & 57.0 & 0.1 & \textbf{69.2} & 1.5 \\
Ours - STD-O~\cite{yu2023spatial}  & 89.8 & 85.7 & 59.4 & \textbf{66.5} & \textbf{52.3} & 76.4 & 89.0 & 90.7 & 65.7 & 71.2 & 62.1 & 69.3 & 75.9 & 65.5 & 63.7 & 18.2 & \textbf{68.8} & 2.8 \\

\hline
\hline
\multicolumn{19}{c}{} \\
\multicolumn{19}{c}{  Plane (PL), Baseball diamond (BD), Bridge (BR), Ground track field (GTF), Small vehicle (SV)} \\
\multicolumn{19}{c}{ Large vehicle (LV), Ship (SH), Tennis court (TC) , Basketball court (BC), Storage tank (ST)} \\
\multicolumn{19}{c}{  Soccer-ball filed (SBF), Roundabout (RA), Harbor (HA), Swimming pool (SP), Helicopter(HC), Container-crane(CC) } \\
\multicolumn{19}{c}{} \\
\end{tabular}
}
\end{center}
\vspace{-2em} 
\end{table*}

\begin{table*}[t]
\centering
\caption{Measured computation complexity of two-stage detectors in GFLOPs.}
\label{Tab:FLOPs}
\begin{tabular}{c|c|c|c|c|c|c|c}
\hline
Model & Backbone & Neck & RPN & High-pass filter & RoI Head &  Total & FPS \\
\hline
Orinted R-CNN~\cite{xie2021oriented}        & 86.1 & 59.4 & 52.0 & - & 13.9 & 211.4 & 19.8 \\
Ours -  Orinted R-CNN~\cite{xie2021oriented}& 86.1 & \textbf{13.4} & \textbf{10.1} & 0.003 & 13.9 & \textbf{123.5} & 30.4 \\
\hline
LSKNet-T~\cite{li2303large}                 & 19.0 & 52.4 & 52.0 & - & 13.9 & 137.3 & 22.3 \\
Ours -  LSKNet-T~\cite{li2303large}         & 19.0 & \textbf{10.2} & \textbf{10.1} & 0.003 &13.9 & \textbf{53.2}   & 36.4 \\
\hline
LSKNet-S~\cite{li2303large}                 & 54.3 & 53.5 & 52.0 & - & 13.9 & 173.7 & 18.2\\
Ours -  LSKNet-S~\cite{li2303large}         & 54.3 & \textbf{10.7}  & \textbf{10.1}& 0.003 & 13.9 & \textbf{89.0}  & 26.3 \\
\hline
STD-HiVit-B~\cite{yu2023spatial}            & 354.7 & 55.3 & 52.0 & - & 1258.3 & 1720.3 & 1.5 \\
Ours - STD-HiVit-B~\cite{yu2023spatial}     & 354.7 & \textbf{11.4} & \textbf{10.1} & 0.003 &\textbf{1249.0} & \textbf{1625.3} & 2.8 \\

\hline
\end{tabular}
\end{table*}

\subsection{Performance evaluation}

Detection accuracy of object classes from various two-stage detectors is shown in Table~\ref{tab:class_mAP}. Models with \lq Ours' in front of the name are the modified models with our method. Last two columns show the mAP and FPS of the model. For ReDet, FPS has increased 1.5$\times$ with 0.2\% accuracy loss with our method. Anchor size was modified as explained in Section~\ref{sec:3.1}, resulting in a loss of accuracy when detecting relatively large objects such as ground track fields and roundabouts, while accuracy for tiny objects like small vehicles and storage tanks improved. A decrease in mAP of 0.4\% is observed with Oriented R-CNN.

When LSKNet-T is used as the backbone, accuracy decreases by 2.1\%. It is observed that accuracy decreases even more when using a high-pass filter.
We are aware that our approach is influenced by the representation power of the backbone.
When using the LSKNet-T backbone, our approach showed low accuracy at baseball fields and harbors. Our high-pass filter increases the score of noisy pixels. A decrease in accuracy of 1.4\% is observed when using LSKNet-S as the backbone. Overall 1.5$\times$ speedup in the model's FPS is observed with our method.

We have profiled computation complexity of implemented two-stage detectors. Table \ref{Tab:FLOPs} shows the FLOPs of each component of the detector. After applying our method, there is no change in FLOPs in the backbone. Because our approach does not contain a feature pyramid, we get reduced FLOPs by 20\% in the neck. Furthermore, by reducing the number of features used as input to the RPN head, there is a 20\% decrease in FLOPs in the RPN. Our method shows a reduction of up to 61.2\% in total FLOPs in the LSKNet-T.

\begin{table}[t]
    \scriptsize
    \centering
    \caption{Performance measurement with various RoIs at the Our-Oriented R-CNN model.}
    \label{Tab:ofRoI}
    \begin{tabular}{l|c|c|c}
    
    \hline
        Model & Total \# of RoIs  &  Accuracy(mAP) & Speed(FPS) \\
        \hline
        Oriented R-CNN(Baseline) & 10,000 &  66.2 & 19.8 \\
        Our-Oriented RCNN        & 10,000 &  66.3 & 22.9 \\
        Our-Oriented RCNN + HPF  & 10,000 &  66.1 & 21.8 \\
        Our-Oriented RCNN        & 6,000  &  66.2 & 26.6 \\
        Our-Oriented RCNN + HPF  & 6,000  &  66.2 & 25.6 \\
        Our-Oriented RCNN        & 2,000  &  65.3 & 30.8 \\                
        Our-Oriented RCNN + HPF  & 2,000  &  65.8 & 30.4 \\
    \hline
    \end{tabular}
\end{table}

We have analyzed the relationship between the number of RoIs and the performance of the model. 
Table \ref{Tab:ofRoI} shows the accuracy and inference speed of our models with varying number of RoIs. A typical two-stage detector has a pyramid network composed of five features, generating 2,000 RoIs from each feature in the feature pyramid, a total of 10,000 RoIs. Our simplified detector uses just one feature, generating 2,000 RoIs, which is five times less than before. The same accuracy of 66.2\% as the baseline model is achieved when using 6,000 RoIs with our HPF filtered model. Our method demonstrates a faster frame rate of 25.6 FPS compared to 19.8 FPS of the baseline with 10,000. Though we further increase RoIs upto 10,000, the accuracy does not increase. This is due to the generation of duplicate RoIs for the same object, resulting in lower IoU between objects and regression boxes after NMS. Our designed high-pass filter is effective to restore accuracy when we reduce the number of RoIs.

\begin{table}[t]
\centering
\caption{Effectiveness of the selected single feature. The accuracy was measured with Our-Oriented R-CNN model.}
\label{Tab:accuracy by feature scale}
\begin{tabular}{c|c|c|c|c|c}
              & P2-P5 & P2 & P3 & P4 & P5\\
\hline
Accuracy(mAP) & 67.5 & 60.9 &  65.3 & 64.0 & 54.4 \\
\hline
\end{tabular}

\end{table}

\begin{table}[t]
    \scriptsize
    \centering
    \caption{Comparison of accuracy before and after using a high-pass filter. All models are applied with 5$\times$5  kernel.}
    \label{Tab:High-pass filter ablation study}
    \begin{tabular}{c|c|c|c}
    \hline
    Model  & High-pass filter & AP & $\Delta$ \\
    \hline
    ReDet~\cite{han2021redet}            &            & 65.5\% & - \\
    ReDet~\cite{han2021redet}             & \checkmark & 66.0\% & +0.5  \\
    Orinted R-CNN~\cite{xie2021oriented}    &            & 65.3\% & -  \\
    Orinted R-CNN~\cite{xie2021oriented}    & \checkmark & 65.8\% & +0.5  \\
    LSKNet-T~\cite{li2303large}         &            & 65.0\% & -  \\
    LSKNet-T~\cite{li2303large}         & \checkmark & 64.2\% & -0.8 \\
    LSKNet-S~\cite{li2303large}         &            & 68.3\% & - \\
    LSKNet-S~\cite{li2303large}         & \checkmark & 68.5\% & +0.2  \\
    STD-HiVit-B~\cite{yu2023spatial}      &            & 68.6\% & -  \\
    STD-HiVit-B~\cite{yu2023spatial}      & \checkmark & 68.8\% & +0.2  \\
    
    \hline
    \end{tabular}
\end{table}

\begin{table*}[ht]
\centering
\caption{Study of various high-pass filters. We used an \textit{unsharp masking filter} in our method.}
\label{Tab:varying high-pass filter}
\begin{tabular}{c|c|c|c|c|c|c|c|c}
\hline
Oriented-RCNN & Baseline & Gaussian & Gaussian & Laplacian & Laplacian & LoG & Ours & Ours \\
\hline
      Filter size    &    -      & 3$\times$3  & 5$\times$5 & 5$\times$5 & 3$\times$3 & 3$\times$3 & 3$\times$3 & 5$\times$5\\
\hline
Accuracy(mAP) & 65.3 & 65.4 &  65.5  & 44.6 & 55.2 & 64.0 & 65.4 & \textbf{65.8}  \\
\hline
\end{tabular}
\end{table*}

\begin{figure*}[t]
    \centering
    \includegraphics[width=1\linewidth]{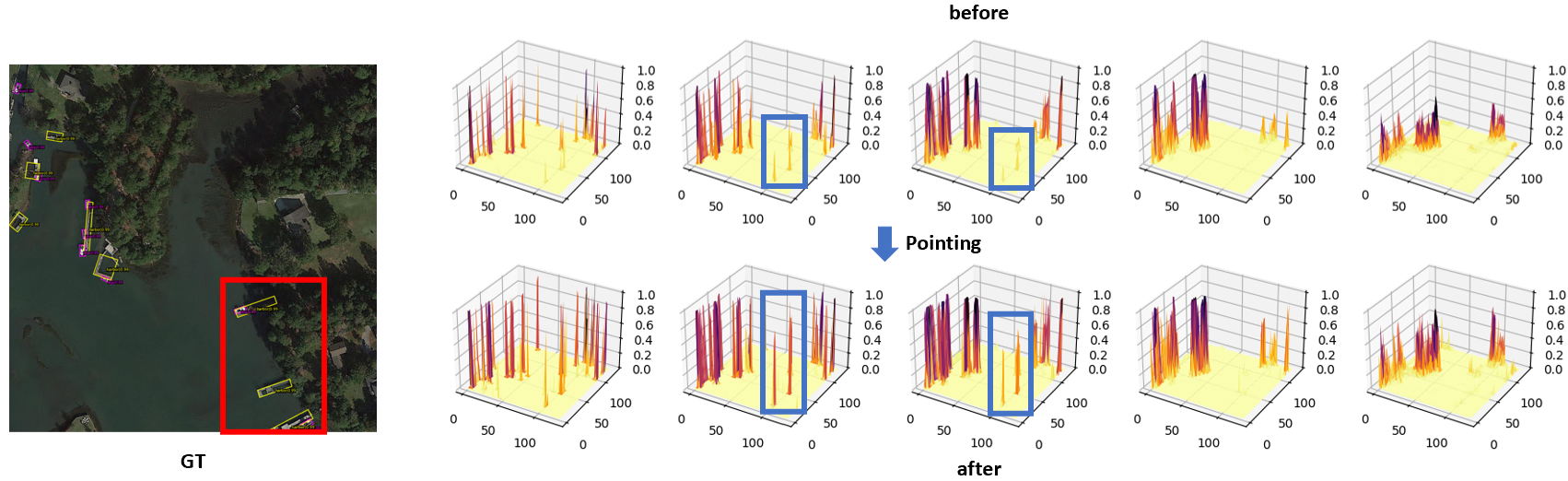}
    \caption{Visualization of the RPN score map before and after applying a high-pass filter. The anchor size increases from left to right. When one axis has sizes of 32 and 64, our method points out the harbor objects with a low score.}
    \label{fig:visualization of the RPN score map}
    
\end{figure*}

\subsection{Analysis of the Selected Feature and HPF Filter Design}

The remaining single feature after removing all the other features from the detector impacts overall accuracy of the simplified network. 
We have measured a detector's accuracy for different scales of features while selecting one feature in the Oriented R-CNN and the result is shown in Table~\ref{Tab:accuracy by feature scale}. The P5 feature has a larger down sample factor than the P2, and anchors become sparse. This makes anchors difficult to detect small objects of DOTA dataset and it has the worst accuracy of 54.4\%. For the P2 feature, the IoU between the object and anchor exceeds the threshold regardless of the anchor's position. However, the selection of the P2 feature leads to the generation of several RoIs from large objects, thereby reducing accuracy. Anchor's sparcity matches with the P3 feature so that it has the highest accuracy of 65.3\%. 

The accuracy before and after applying a high-pass filter to the model using our method is shown in Table~\ref{Tab:High-pass filter ablation study}. Improvements in accuracy can be observed across various models. However, the accuracy of the LSKNet-T model decreases. Our high-pass filter enhances accuracy for objects with rectangular shapes, such as ships and bridges while maintaining accuracy for large objects. However, with LSKNet-T, there is a decrease in accuracy at the baseball diamond and harbor class. 
Figure~\ref{fig:visualization of the RPN score map} provides a visual demonstration of how our high-pass filter effectively improves the accuracy. The area corresponding to the red box in Figure \ref{fig:visualization of the RPN score map} has a ship and a harbor. However, the score of the object in that area is low. Our high-pass filter sharpens the score to detect objects.(Blue box). However, it can be seen that the score increases even in areas unrelated to the object.

We compare the results of applying several high-pass filters in Table~\ref{Tab:varying high-pass filter}. Even when using a Gaussian filter, a slight improvement in accuracy is observed. The Laplacian filter results in a significant decrease in accuracy. The Laplacian of Gaussian hybrid method achieved an accuracy of 65.4\%. The unsharp masking filter in our approach with 5$\times$5 kernel size yielded the most significant improvement, with an accuracy increase of 0.5\% compared to the baseline.

\section{Conclusion}
\label{sec:conc}

In this paper, we study methods to simplify models for real-time on-board inference from remote sensing images.
We reduce computational overhead by not constructing a feature pyramid and instead using a single feature at two-stage detectors that achieve state-of-the-art accuracy. 
Using a single feature results in a loss of accuracy. We propose a method for selecting features that minimizes loss of accuracy.
Additionally, we adjust the anchor size to fit the dataset to recover lost accuracy. To avoid the concentration of RoIs being generated on large objects, a high-pass filter is included.
Our method reduces FLOPs by 61.2\% with a 2.1\% decrease in accuracy in the LSKNet-T model. Similar results are seen in other two-stage models.
However, we have limitations in accuracy loss compared to the baseline. Furthermore, the high-pass filter increases the score even in areas where the item does not really exist, leading to the generation of noise.
Our approach was evaluated only in the two-stage detector. Research on methods that can also be used in one-stage detectors is needed.
We need further research on methods to recover or enhance accuracy with a slight increase in computational workload.

\bibliographystyle{unsrt}
\bibliography{bibliography}

\begin{thebibliography}{10}

\bibitem{qi2018board}
Baogui Qi, Hao Shi, Yin Zhuang, He~Chen, and Liang Chen.
\newblock On-board, real-time preprocessing system for optical remote-sensing imagery.
\newblock {\em Sensors}, 18(5):1328, 2018.

\bibitem{shen2023board}
Yanyun Shen, Di~Liu, Junyi Chen, Zhipan Wang, Zhe Wang, and Qingling Zhang.
\newblock On-board multi-class geospatial object detection based on convolutional neural network for high resolution remote sensing images.
\newblock {\em Remote Sensing}, 15(16):3963, 2023.

\bibitem{ghiglione2022opportunities}
Max Ghiglione and Vittorio Serra.
\newblock Opportunities and challenges of ai on satellite processing units.
\newblock In {\em Proceedings of the 19th ACM international conference on computing Frontiers}, pages 221--224, 2022.

\bibitem{li2303large}
Y~Li, Q~Hou, Z~Zheng, MM~Cheng, J~Yang, and X~Li.
\newblock Large selective kernel network for remote sensing object detection. arxiv 2023.
\newblock {\em arXiv preprint arXiv:2303.09030}.

\bibitem{yu2023spatial}
Hongtian Yu, Yunjie Tian, Qixiang Ye, and Yunfan Liu.
\newblock Spatial transform decoupling for oriented object detection.
\newblock {\em arXiv preprint arXiv:2308.10561}, 2023.

\bibitem{xie2021oriented}
Xingxing Xie, Gong Cheng, Jiabao Wang, Xiwen Yao, and Junwei Han.
\newblock Oriented r-cnn for object detection.
\newblock In {\em Proceedings of the IEEE/CVF international conference on computer vision}, pages 3520--3529, 2021.

\bibitem{zhao2019robust}
Yiming Zhao, Jinzheng Zhao, Chunyu Zhao, Weiyu Xiong, Qingli Li, and Junli Yang.
\newblock Robust real-time object detection based on deep learning for very high resolution remote sensing images.
\newblock In {\em IGARSS 2019-2019 IEEE International Geoscience and Remote Sensing Symposium}, pages 1314--1317. IEEE, 2019.

\bibitem{lyu2023survey}
Zonglei Lyu, Tong Yu, Fuxi Pan, Yilin Zhang, Jia Luo, Dan Zhang, Yiren Chen, Bo~Zhang, and Guangyao Li.
\newblock A survey of model compression strategies for object detection.
\newblock {\em Multimedia Tools and Applications}, pages 1--72, 2023.

\bibitem{chen2021you}
Qiang Chen, Yingming Wang, Tong Yang, Xiangyu Zhang, Jian Cheng, and Jian Sun.
\newblock You only look one-level feature.
\newblock In {\em Proceedings of the IEEE/CVF conference on computer vision and pattern recognition}, pages 13039--13048, 2021.

\bibitem{yi2023point2rbox}
Yu~Yi, Xue Yang, Qingyun Li, Feipeng Da, Junchi Yan, Jifeng Dai, and Yu~Qiao.
\newblock Point2rbox: Combine knowledge from synthetic visual patterns for end-to-end oriented object detection with single point supervision.
\newblock {\em arXiv preprint arXiv:2311.14758}, 2023.

\bibitem{xia2018dota}
Gui-Song Xia, Xiang Bai, Jian Ding, Zhen Zhu, Serge Belongie, Jiebo Luo, Mihai Datcu, Marcello Pelillo, and Liangpei Zhang.
\newblock Dota: A large-scale dataset for object detection in aerial images.
\newblock In {\em Proceedings of the IEEE conference on computer vision and pattern recognition}, pages 3974--3983, 2018.

\bibitem{lin2017focal}
Tsung-Yi Lin, Priya Goyal, Ross Girshick, Kaiming He, and Piotr Doll{\'a}r.
\newblock Focal loss for dense object detection.
\newblock In {\em Proceedings of the IEEE international conference on computer vision}, pages 2980--2988, 2017.

\bibitem{han2021redet}
Jiaming Han, Jian Ding, Nan Xue, and Gui-Song Xia.
\newblock Redet: A rotation-equivariant detector for aerial object detection.
\newblock In {\em Proceedings of the IEEE/CVF Conference on Computer Vision and Pattern Recognition}, pages 2786--2795, 2021.

\bibitem{zhou2021probabilistic}
Xingyi Zhou, Vladlen Koltun, and Philipp Kr{\"a}henb{\"u}hl.
\newblock Probabilistic two-stage detection.
\newblock {\em arXiv preprint arXiv:2103.07461}, 2021.

\bibitem{terven2023comprehensive}
Juan Terven and Diana Cordova-Esparza.
\newblock A comprehensive review of yolo: From yolov1 to yolov8 and beyond.
\newblock {\em arXiv preprint arXiv:2304.00501}, 2023.

\bibitem{dai2021dynamic}
Xiyang Dai, Yinpeng Chen, Bin Xiao, Dongdong Chen, Mengchen Liu, Lu~Yuan, and Lei Zhang.
\newblock Dynamic head: Unifying object detection heads with attentions.
\newblock In {\em Proceedings of the IEEE/CVF conference on computer vision and pattern recognition}, pages 7373--7382, 2021.

\bibitem{ma2018arbitrary}
Jianqi Ma, Weiyuan Shao, Hao Ye, Li~Wang, Hong Wang, Yingbin Zheng, and Xiangyang Xue.
\newblock Arbitrary-oriented scene text detection via rotation proposals.
\newblock {\em IEEE transactions on multimedia}, 20(11):3111--3122, 2018.

\bibitem{ding2019learning}
Jian Ding, Nan Xue, Yang Long, Gui-Song Xia, and Qikai Lu.
\newblock Learning roi transformer for oriented object detection in aerial images.
\newblock In {\em Proceedings of the IEEE/CVF Conference on Computer Vision and Pattern Recognition}, pages 2849--2858, 2019.

\bibitem{lyu2022rtmdet}
Chengqi Lyu, Wenwei Zhang, Haian Huang, Yue Zhou, Yudong Wang, Yanyi Liu, Shilong Zhang, and Kai Chen.
\newblock Rtmdet: An empirical study of designing real-time object detectors.
\newblock {\em arXiv preprint arXiv:2212.07784}, 2022.

\bibitem{girshick2014rich}
Ross Girshick, Jeff Donahue, Trevor Darrell, and Jitendra Malik.
\newblock Rich feature hierarchies for accurate object detection and semantic segmentation.
\newblock In {\em Proceedings of the IEEE conference on computer vision and pattern recognition}, pages 580--587, 2014.

\bibitem{girshick2015fast}
Ross Girshick.
\newblock Fast r-cnn.
\newblock In {\em Proceedings of the IEEE international conference on computer vision}, pages 1440--1448, 2015.

\bibitem{ren2015faster}
Shaoqing Ren, Kaiming He, Ross Girshick, and Jian Sun.
\newblock Faster r-cnn: Towards real-time object detection with region proposal networks.
\newblock {\em Advances in neural information processing systems}, 28, 2015.

\bibitem{redmon2016you}
Joseph Redmon, Santosh Divvala, Ross Girshick, and Ali Farhadi.
\newblock You only look once: Unified, real-time object detection.
\newblock In {\em Proceedings of the IEEE conference on computer vision and pattern recognition}, pages 779--788, 2016.

\bibitem{redmon2017yolo9000}
Joseph Redmon and Ali Farhadi.
\newblock Yolo9000: better, faster, stronger.
\newblock In {\em Proceedings of the IEEE conference on computer vision and pattern recognition}, pages 7263--7271, 2017.

\bibitem{lin2017feature}
Tsung-Yi Lin, Piotr Doll{\'a}r, Ross Girshick, Kaiming He, Bharath Hariharan, and Serge Belongie.
\newblock Feature pyramid networks for object detection.
\newblock In {\em Proceedings of the IEEE conference on computer vision and pattern recognition}, pages 2117--2125, 2017.

\bibitem{zhou2019objects}
Xingyi Zhou, Dequan Wang, and Philipp Kr{\"a}henb{\"u}hl.
\newblock Objects as points.
\newblock {\em arXiv preprint arXiv:1904.07850}, 2019.

\bibitem{zhou2022mmrotate}
Yue Zhou, Xue Yang, Gefan Zhang, Jiabao Wang, Yanyi Liu, Liping Hou, Xue Jiang, Xingzhao Liu, Junchi Yan, Chengqi Lyu, Wenwei Zhang, and Kai Chen.
\newblock Mmrotate: A rotated object detection benchmark using pytorch.
\newblock In {\em Proceedings of the 30th ACM International Conference on Multimedia}, 2022.

\end{thebibliography}

\clearpage

\end{document}